\documentclass[twoside]{article}

\usepackage{amsmath,amssymb}
\usepackage{graphicx}
\usepackage{url}
\usepackage{algorithmicx,algpseudocode,algorithm}

\usepackage{enumitem}
\setlist{itemsep=-1pt,topsep=.6ex,leftmargin=4ex} 

\def\epsilon{\varepsilon}
\def\phi{\varphi}
\def\<{\langle}
\def\>{\rangle}
\def\...{,\ldots,}

\def\e{{\rm e}}
\def\refeq#1{{(\ref{eq:#1})}}
\def\argmax{\mathop{\rm argmax}}

\def\given{{\,|\,}}
\def\conv{\mathop{\rm conv}}

\def\bb#1{\mathbb{#1}} 

\usepackage[amsmath,thmmarks]{ntheorem}
\theorembodyfont{\slshape}
\theoremseparator{.}
\theoremsymbol{}

\newtheorem{theorem}{Theorem}

\theorembodyfont{\normalfont}
\theoremsymbol{\mbox{$\square$}}

\theoremheaderfont{\it}
\newtheorem*{proof}{Proof}

\usepackage[rreport]{cmpcover}
\title{Zero-Temperature Limit of a Convergent Algorithm to Minimize the Bethe Free Energy}
\author{Tom{\'a}{\v s} Werner}
\date{December 2011}
\CMPReportNo{CTU--CMP--2011--14}
\CMPAcknowledgement{The work has been supported by the European Commission
project FP7-ICT-270138 and the Czech Grant Agency project P103/10/0783.}
\CMPDocumentURL{ftp://cmp.felk.cvut.cz/pub/cmp/articles/werner/Werner-TR-2011-14.pdf}

\begin{document}

\maketitle

\begin{abstract}
\noindent
After the discovery that fixed points of loopy belief propagation
coincide with stationary points of the Bethe free energy, several
researchers proposed provably convergent algorithms to directly
minimize the Bethe free energy. These algorithms were formulated only
for non-zero temperature (thus finding fixed points of the sum-product
algorithm) and their possible extension to zero temperature is not
obvious. We present the zero-temperature limit of the double-loop
algorithm by Heskes, which converges a max-product fixed point. The
inner loop of this algorithm is max-sum diffusion. Under certain
conditions, the algorithm combines the complementary advantages of the
max-product belief propagation and max-sum diffusion (LP relaxation):
it yields good approximation of both ground states and max-marginals.
\end{abstract}

\section{Introduction}

Loopy belief propagation \cite{Pearl88} is a well-known algorithm to
approximate marginals of the Gibbs distribution defined by an
undirected graphical model. For acyclic graphs, BP always converges
and yields the exact marginals. For graphs with cycles, it is not
guaranteed to converge but when it does, it often yields surprisingly
good approximations of the true marginals. One informal argument for
this is that at a BP fixed point, marginals are exact in every
sub-tree of the factor graph \cite{Wainwright03d,Wainwright04}.
Attempts to understand loopy BP has generated a large body of
literature, see e.g.\ the survey \cite{Wainwright08}.

BP has a modification, known as the max-product BP, where summations
are replaced with maximizations. In statistical mechanics terminology,
this can be understood as the zero-temperature limit of the ordinary BP.
Max-product BP computes (or approximates) max-marginals rather than
ordinary marginals.

After the discovery \cite{Yedidia00,Yedidia05} that BP fixed points
coincide with stationary points of the Bethe free energy, several
researchers proposed provably convergent algorithms to find a local
minimum of the Bethe free energy
\cite{Yuille-NC02,Welling-UAI-2001,Teh-NIPS-2001,Heskes-NIPS-2003,Heskes-JAIR-2006}.
These algorithms have been proposed only for the sum-product and their
possible extension to the max-product is not obvious.

We reformulate the double-loop algorithm \cite{Heskes-NIPS-2003} by
Heskes such that taking its zero-temperature limit becomes
straightforward, which results in an algorithm that always converges
to a max-product BP fixed point. The inner loop of the algorithm is
max-sum diffusion
\cite{Kovalevsky-diffusion,Werner-PAMI07,Werner-PAMI-2010,Franc-MITbook-2011}.
We empirically observed that with a uniform initialization, the
algorithm always yielded the same approximation of ground states that
would be obtained by max-sum diffusion (or other algorithms for MAP
inference based on LP relaxation, such as TRW-S \cite{Kolmogorov06}).
Thus, it combines the complementary advantages of max-sum belief
propagation and LP relaxation: unlike the former, it yields good
approximation of ground states and, unlike the latter, it yields a
good approximation of max-marginals.

The text is organized as follows. We first (\S\ref{sec:gibbs}) review
the basics of inference in graphical models. We thoroughly discuss the
zero-temperature limit of the Gibbs distribution and related
quantities and how to obtain their approximation by variational
inference.  Then we review two basic cases of variational inference,
with a convex free energy (\S\ref{sec:diffusion}) and with the Bethe
free energy (\S\ref{sec:BP}). Then (\S\ref{sec:diffusion:inf},
\S\ref{sec:BP:inf}) we discuss their zero-temperature limits in
detail. Finally (\S\ref{sec:double-loop}) we reformulate the
double-loop algorithm \cite{Heskes-NIPS-2003} and modify it for the
zero temperature.


\section{Gibbs distribution}
\label{sec:gibbs}

Let $V$ be a set of variables, each variable $v\in V$ taking states
$x_v$ from a finite domain $X_v$. An assignment to a variable subset
$a\subseteq V$ is $x_a\in X_a$, where $X_a$ is the Cartesian product
of domains $X_v$ for $v\in a$. In particular, $x_V\in X_V$ is an
assignment to all the variables.  Let $E\subseteq 2^V$, thus $(V,E)$
is a hypergraph.  Each variable $v\in V$ and hyperedge $a\in E$ is
assigned a potential function $\theta_v{:}\ X_v\to\overline{\bb R}$
and $\theta_a{:}\ X_a\to\overline{\bb R}$, respectively, where
$\overline{\bb R}=\bb R\cup\{-\infty\}$. All numbers $\theta_v(x_v)$
and $\theta_a(x_a)$ are understood as a single vector
$\theta\in\overline{\bb R}^I$ (or mapping $\theta{:}\
I\to\overline{\bb R}$) with
\[
I = \{\, (v,x_v) \mid v\in V, \; x_v\in X_v \,\} \cup \{\, (a,x_a)
\mid a\in E, \; x_a\in X_a \,\} .
\]
The Gibbs probability distribution over the hypergraph $(V,E)$ is given by
\begin{equation}
\label{eq:p}
p(x_V) = \exp[\, \<\theta,\delta(x_V)\> - \Phi(\theta) \,]
\end{equation}
where the mapping $\delta{:}\ X_V\to\{0,1\}^I$ is such that
\begin{equation}
\<\theta,\delta(x_V)\> =
\sum_{v\in V} \theta_v(x_v) +
\sum_{a\in E} \theta_a(x_a) .
\label{eq:Gibbs}
\end{equation}
For infinite weights, we set $-\infty\cdot0=0$ in the scalar product
$\<\theta,\delta(x_V)\>$. Since unary terms are included
in~\refeq{Gibbs} explicitly, we assume that $E$ contains no
singletons. The distribution is normalized by the {\em log-partition
  function\/}
\begin{equation}
\Phi(\theta)
= \log \sum_{x_V} \exp\<\theta,\delta(x_V)\>
= \bigoplus_{x_V} \<\theta,\delta(x_V)\> .
\label{eq:log-partition}
\end{equation}

In~\refeq{log-partition}, we used $x \oplus y =
\log(\e^x+\e^y)$ to denote the {\em log-sum-exp operation\/}. It will
be useful to keep in mind algebraic properties of this operation. It
is associative and commutative, and addition distributes over it.
Thus, $(\overline{\bb R},\oplus,+)$ is a commutative semiring. This
semiring is, via the logarithm map, isomorphic to the `sum-product'
semiring $(\bb R_+,+,\times)$.

\paragraph{Marginals.}
The marginals of the distribution are
\begin{equation}
\mu_v(x_v) = \sum_{x_{V\setminus v}} p(x_V) , \quad
\mu_a(x_a) = \sum_{x_{V\setminus a}} p(x_V) ,
\label{eq:marginals}
\end{equation}
where we abuse notation by writing $V\setminus v$ instead of
$V\setminus\{v\}$. The numbers~\refeq{marginals} are understood as a
vector $\mu\in[0,1]^I$. All realizable marginal vectors $\mu$ form the
{\em marginal polytope\/} $\conv\delta(X_V)$, where $\delta(X_V)=\{\,
\delta(x_V) \mid x_V\in X_V \,\}$. Besides (an exponential number of)
other constraints, $\mu$~satisfies normalization and marginalization
constraints
\begin{equation}
\label{eq:local-polytope}
\sum_{x_v} \mu_v(x_v) = \sum_{x_a} \mu_a(x_a) = 1 , \quad
\sum_{x_{a\setminus v}} \mu_a(x_a) = \mu_v(x_v) .
\end{equation}
All vectors $\mu\ge0$ satisfying~\refeq{local-polytope} form the {\em
  local marginal polytope\/} $\Lambda$. We have
$\conv\delta(X_V)\subseteq\Lambda$, with equality if and only if
hypergraph $(V,E)$ is acyclic (i.e., its factor graph is a tree).  We
also introduce a symbol for log-marginals,
\begin{equation}
\nu_a(x_a) = \log\mu_a(x_a) = \bigoplus_{x_{V\setminus a}} \<\theta,\delta(x_V)\> - \Phi(\theta)
\label{eq:log-marginals}
\end{equation}
(and similarly for $\nu_v(x_v)$). For log-marginals, 
constraints~\refeq{local-polytope} read
\begin{equation}
\label{eq:log-local-polytope}
\bigoplus_{x_v} \nu_v(x_v) = \bigoplus_{x_a} \nu_a(x_a) = 0 , \quad
\bigoplus_{x_{a\setminus v}} \nu_a(x_a) = \nu_v(x_v) .
\end{equation}

\paragraph{Reparameterizations.}
A {\em reparameterization\/} is an affine transformation of vector
$\theta$ that preserves~\refeq{Gibbs} for all assignments $x_V\in
X_V$. We first define the {\em local reparameterization\/} on a pair
$(a,v)$ as follows: subtract an arbitrary unary function $\alpha_{av}{:}\ X_v\to\bb
R$ from $\theta_v$ and add the same function to $\theta_a$,
\begin{equation}
\theta_v(x_v) \gets \theta_v(x_v) - \alpha_{av}(x_v) , \quad
\theta_a(x_a) \gets \theta_a(x_a) + \alpha_{av}(x_v) .
\label{eq:LET}
\end{equation}
This preserves~\refeq{Gibbs} because $\alpha_{av}$ cancels out.  We
understand~\refeq{LET} as `passing a message' $\alpha_{av}$. Applying
local reparameterization~\refeq{LET} to all pairs $(a,v)$ with $v\in
a\in E$ yields the general reparameterization
\begin{equation}
\theta^\alpha_v(x_v) = \theta_v(x_v) - \sum_{a \ni v} \alpha_{av}(x_v) , \quad
\theta^\alpha_a(x_a) = \theta_a(x_a) + \sum_{v \in a} \alpha_{av}(x_v)
\label{eq:eqt}
\end{equation}
where $\alpha=\{\,\alpha_{av}(x_v) \mid v\in a\in E, \; x_v\in X_v
\,\}$ is the vector of all messages and the transformed vector
$\theta$ is denoted $\theta^\alpha\in\overline{\bb R}^I$. Thus
$\<\theta^\alpha,\delta(x_V)\>=\<\theta,\delta(x_V)\>$. In fact, we
have more generally $\<\theta^\alpha,\mu\>=\<\theta,\mu\>$ for all
$\mu$ satisfying~\refeq{local-polytope} and all $\alpha$.

Reparameterizations can be done either by directly modifying the
vector $\theta$ or by keeping $\theta$ unchanged and storing the
messages $\alpha$. While the former may be better for theoretical
analysis, the latter is preferable in algorithms. In the sequel we
freely switch between these two views.

\subsection{Zero-temperature limit}
\label{sec:gibbs:inf}

In this section, we will use $p(x_V\given \theta)$ and
$\mu_v(x_v\given \theta)$, $\mu_a(x_a\given \theta)$ to explicitly
denote the dependence of distribution~\refeq{p} and its marginals
on~$\theta$. 

In statistical physics, the Gibbs distribution is usually considered in a
more general form as $p(x_V\given\beta \theta)$, where $\beta>0$ is
the inverse temperature \cite{Mezard-Montanari2009}. The limit
$\beta\to\infty$ is then known as the {\em zero-temperature limit\/}.

It is elementary to show that the distribution
\begin{equation}
p^\infty(x_V\given \theta) = \lim_{\beta\to\infty} p(x_V\given\beta \theta)
\label{eq:p-infty}
\end{equation}
is zero everywhere except at {\em ground states\/}, which are
the maximizers of $p(x_V\given \theta)$ or, equivalently,
$\<\theta,\delta(x_V)\>$. If there are multiple ground states then the
mass is distributed evenly among them.

The zero-temperature limit of the log-partition function is
\begin{equation}
\Phi^\infty(\theta)
= \lim_{\beta\to\infty} \frac{\Phi(\beta \theta)}{\beta}
= \max_{x_V} \<\theta,\delta(x_V)\> ,
\label{eq:log-partition-infty}
\end{equation}
which follows from the limit
\begin{equation}
\lim_{\beta\to\infty} \frac{(\beta x)\oplus(\beta y)}{\beta} = \max\{x,y\} .
\label{eq:oplus-max}
\end{equation}
The zero-temperature limit of log-marginals~\refeq{log-marginals}
yields {\em max-marginals\/}\footnote{
  It would be more precise to call~\refeq{max-marginals}
  `max-log-marginals' or `log-max-marginals'. But, as is usual in the
  literature, we call them only `max-marginals'.
}
\begin{equation}
\nu^\infty_a(x_a)
= \lim_{\beta\to\infty} \frac{\nu_a(x_a\given\beta \theta)}{\beta}
= \max_{x_{V\setminus a}} \<\theta,\delta(x_V)\> - \Phi^\infty(\theta)
\label{eq:max-marginals}
\end{equation}
(similarly for $\nu^\infty_v(x_v)$).  Observe
that~\refeq{max-marginals} and~\refeq{log-partition-infty} differs
from~\refeq{log-marginals} and~\refeq{log-partition} only by replacing
the log-sum-exp operation `$\oplus$' with `$\max$'.  This corresponds,
by the limit~\refeq{oplus-max}, to transition from the semiring
$(\overline{\bb R},\oplus,+)$ to the max-sum semiring $(\overline{\bb
  R},\max,+)$. Similarly, max-marginals satisfy normalization and
marginalization conditions \refeq{log-local-polytope} in which
`$\oplus$' has been replaced with `$\max$'.

Max-marginals should not be confused\footnote{
  Unlike the limit~\refeq{p-infty}, the limit~\refeq{max-marginals}
  from (log-)marginals to max-marginals rarely appears in the machine
  learning or computer vision literature. The only work we found is
  \cite{Johnson-SSG-2005}.
} with the marginals of $p^\infty(x_V\given \theta)$. These are
different quantities and one cannot be computed from the other.

\paragraph{Recovering ground states from max-marginals.}
Ground states can be recovered from max-marginals. To show that, we
first recall what is the {\em constraint satisfaction problem \/}
(CSP) \cite{Mackworth91,Cohen06}. The CSP instance is defined by a
vector $c\in\{0,1\}^I$, where functions $c_v{:}\ X_v\to\{0,1\}$ and
$c_a{:}\ X_a\to\{0,1\}$ are understood as relations. A solution of the
CSP is an assignment $x_V$ satisfying all the relations, i.e.,
$c_v(x_v)=1$ for all $v\in V$ and $c_a(x_a)=1$ for all $a\in E$.

For a vector $\theta\in\overline{\bb R}^I$ we define vector
$\lceil\theta\rceil\in\{0,1\}^I$ by
\[
\lceil \theta\rceil_v(x_v) =
\begin{cases}
1 & \text{if}\; \displaystyle x_v\in\argmax_{y_v}\theta_v(y_v) \\
0 & \text{otherwise}
\end{cases} , \;\;
\lceil \theta\rceil_a(x_a) =
\begin{cases}
1 & \text{if}\; \displaystyle x_a\in\argmax_{y_a}\theta_a(y_a) \\
0 & \text{otherwise}
\end{cases} ,
\]
i.e., a component of $\lceil\theta\rceil$ equals $1$ iff the
corresponding component of $\theta$ is maximal in its potential
function. We say that such a components of $\theta$ is {\em active\/}.
Now the set $\argmax_{x_V}\<\theta,\delta(x_V)\>$ of ground states is
the solution set of the CSP defined by vector $\lceil
\nu^\infty\rceil$ of active max-marginals.

\subsection{Convex conjugacy and variational inference}
\label{sec:var-inference}

Let $H(\mu)$ denote the entropy of the distribution \refeq{p} as a
function of its marginals. The functions $\Phi(\theta)$ and $-H(\mu)$
are convex and they are related by convex conjugacy,
\begin{equation}
\Phi(\theta) = \max_{\mu\in\conv\delta(X_V)} [ \<\theta,\mu\> + H(\mu) ] ,
\label{eq:conjugacy}
\end{equation}
where the optimum is attained for $\mu$ equal to the marginals
\refeq{marginals}. In statistical physics, the quantity
$-\<\theta,\mu\>-H(\mu)$ is known as the {\em Gibbs free energy\/} of
the system.  By taking the limit $\beta\to\infty$ of the expression
\begin{equation}
\frac{\Phi(\beta\theta)}{\beta} =
\max_{\mu\in\conv\delta(X_V)} \bigg[ \<\theta,\mu\> + \frac{H(\mu)}{\beta} \bigg]
\label{eq:conjugacy:beta}
\end{equation}
we similarly obtain $\Phi^\infty(\theta)$ and max-marginals
$\nu^\infty$.

The idea behind {\em variational inference\/}~\cite{Wainwright08} is
to replace the marginal polytope $\conv\delta(X_V)$ and the entropy
$H(\mu)$ in~\refeq{conjugacy} with their tractable approximations.
Then the optimal value and the optimal argument of~\refeq{conjugacy}
is an approximation of the log-partition function and marginals,
respectively. For $\beta\to\infty$,
\begin{itemize}
\item the optimal value of~\refeq{conjugacy:beta} is an approximation
of $\Phi^\infty(\theta)$,
\item the logarithm of the optimal argument of~\refeq{conjugacy:beta}
is an approximation of max-marginals $\nu^\infty$,
\item the solution set of the CSP defined by active approximate
max-marginals is an approximation of the set
$\argmax_{x_V}\<\theta,\delta(x_V)\>$ of ground states.
\end{itemize}
As the entropy term in~\refeq{conjugacy:beta} approaches $0$ for
$\beta\to\infty$, one may think that it could be simply omitted.
However, as pointed out in~\cite{Weiss-UAI07}, if the approximate
entropy is non-convex (such as the Bethe entropy), the
problem~\refeq{conjugacy:beta} can have multiple local minima for
arbitrarily large $\beta$. Thus, if our algorithm finds only a local
minimum of~\refeq{conjugacy:beta}, the entropy term is crucial.


\section{Convex free energy}
\label{sec:diffusion}

Let the marginal polytope in~\refeq{conjugacy} be approximated by the
local marginal polytope $\Lambda$ and the true entropy by
$H(\mu)\approx-\<\log\mu,\mu\>$. This entropy approximation is
concave, thus we obtained a simple (arguably, the simplest possible)
variational inference method with a convex free energy
\cite{Weiss-UAI07,Hazan-UAI08}. The approximation of~\refeq{conjugacy}
now reads
\begin{equation}
\max_{\mu\in \Lambda} \, \langle \theta-\log\mu,\mu\rangle .
\label{eq:primal-convex}
\end{equation}

The problem~\refeq{primal-convex} can be solved as described e.g.\ in
\cite{Werner-CVWW07}. Its dual reads as follows: find a
reparameterization of the original vector $\theta$ that minimizes the
function
\begin{equation}
U(\theta) = \sum_{v\in V} \bigoplus_{x_v} \theta_v(x_v) + \sum_{a\in E} \bigoplus_{x_a} \theta_a(x_a) .
\label{eq:U}
\end{equation}
This is a majorant of the log-partition function,
$U(\theta)\ge\Phi(\theta)$ for every $\theta$.
A~sufficient condition for dual optimality is that
\begin{equation}
\bigoplus_{x_{a\setminus v}} \theta_a(x_a) = \theta_v(x_v)
\label{eq:fixed:diffusion}
\end{equation}
for all $v\in a\in E$ and $x_v\in X_v$.
The primal and dual optimum are related by
\begin{equation}
\log\mu_v(x_v) = \theta_v(x_v) - \bigoplus_{y_v} \theta_v(y_v) , \quad
\log\mu_a(x_a) = \theta_a(x_a) - \bigoplus_{y_a} \theta_a(y_a) .
\label{eq:marginals:diffusion}
\end{equation}

Since function \refeq{U} is convex and differentiable, its global
minimum over reparameterizations of $\theta$ can be found by
coordinate descent. This leads to a simple message passing algorithm.
The iteration of this algorithm enforces equality
\refeq{fixed:diffusion} for a single pair $(a,v)$ by local
reparameterization~\refeq{LET}, which determines~$\alpha_{av}(x_v)$
in~\refeq{LET} uniquely. The iteration decreases $U(\theta)$, and this
decrease is strict unless $U(\theta)$ is already minimal.  On
convergence, \refeq{fixed:diffusion} holds globally.

If reparameterizations are represented by messages rather than by
directly modifying $\theta$, the dual of \refeq{primal-convex} reads
$\min_\alpha U(\theta^\alpha)$ and the coordinate descent becomes
Algorithm~\ref{alg:diffusion}.  To correctly handle infinite weights,
the algorithm expects that
$[\theta_v(x_v)>-\infty]\Leftrightarrow[\max_{x_{a\setminus v}}
\theta_a(x_a)>-\infty]$ for all $v\in a\in E$ and $x_v\in X_v$.

\begin{algorithm}[htb]
\caption{The `diffusion' algorithm.}
\label{alg:diffusion}
\begin{algorithmic}
\Repeat
\For { $v\in a\in E$ and $x_v\in X_v$ such that $\theta_v(x_v)>-\infty$ }
  \State $\displaystyle \alpha_{av}(x_v) \gets \alpha_{av}(x_v) + {1\over2} \Big[ \theta^\alpha_v(x_v) - \bigoplus_{x_{a\setminus v}} \theta^\alpha_a(x_a) \Big]$
\EndFor
\Until { convergence }
\end{algorithmic}
\end{algorithm}

\subsection{Zero-temperature limit: max-sum diffusion}
\label{sec:diffusion:inf}

The zero-temperature limit of the optimization problem above is
obtained by replacing $\theta$ with $\beta \theta$ and taking the
limit $\beta\to\infty$. This results in replacing `$\oplus$' with
`$\max$' in~\refeq{U}--\refeq{marginals:diffusion} and 
Algorithm~\ref{alg:diffusion}. We assume that this has been done.

This yields the LP relaxation approach to maximizing the Gibbs
distribution first proposed by Schlesinger et al.\
\cite{Schlesinger76,Kovalevsky-diffusion}, see also
\cite{Werner-PAMI07,Werner-CVWW07,Werner-PAMI-2010,Franc-MITbook-2011}.
In these works, the zero-temperature limit of
Algorithm~\ref{alg:diffusion} is called {\em max-sum diffusion\/}.

Let function~\refeq{U} after replacing
`$\oplus$' with `$\max$' be denoted by
\[
U^\infty(\theta) =
\lim_{\beta\to\infty} \frac{U(\beta\theta)}{\beta} =
\sum_{v\in V} \max_{x_v} \theta_v(x_v) + \sum_{a\in E} \max_{x_a} \theta_a(x_a) .
\]
We have $U^\infty(\theta)\ge \Phi^\infty(\theta)$ for every $\theta$.
Algorithm~\ref{alg:diffusion} tries to minimize $U^\infty(\theta)$ by
reparameterizing $\theta$.  However, the function $U^\infty$ is
non-differentiable now -- therefore Algorithm~\ref{alg:diffusion} may
converge only to a local (with respect to coordinate moves) minimum of
$U^\infty(\theta)$. While it is easy to prove convergence of the
algorithm in value, convergence in argument is only a conjecture to
date \cite{Werner-PAMI07,Werner-PAMI-2010} and only a weaker property
has been proved recently \cite{Schlesinger-2011}.


According to~\S\ref{sec:var-inference}, when $\theta$ is optimal then
$U^\infty(\theta)$ is an approximation of $\Phi^\infty(\theta)$
and~\refeq{marginals:diffusion} is an approximation of the
max-marginals~$\nu^\infty$. Note that the approximate
max-marginals~\refeq{marginals:diffusion} are, up to normalization,
directly equal to~$\theta$. Since vector $\lceil\theta\rceil$ is not
affected by normalization of $\theta$, the solution set of the
CSP~$\lceil\theta\rceil$ is an approximation of the ground states.

But this is in agreement with \cite{Schlesinger76,Werner-PAMI07},
where it is shown that the inequality $U^\infty(\theta)\ge
\Phi^\infty(\theta)$ (and hence the LP relaxation) is tight if and
only if the CSP defined by $\lceil\theta\rceil$ has a solution.  Then,
$\<\theta,\delta(x_V)\>=\Phi^\infty(\theta)$ for every solution $x_V$
of CSP $\lceil\theta\rceil$. There are two important problem
subclasses for which the LP relaxation is tight: if hypergraph $(V,E)$
is acyclic or if the functions $\theta_a$ are (permuted) supermodular
\cite{Werner-PAMI07,Werner-PAMI-2010,Kolmogorov05}.  Besides, it is
tight for many other instances met in applications. This makes this
method very suitable for approximating ground states, which has been
also observed empirically\footnote{
  The TRW-S algorithm \cite{Kolmogorov06} studied in \cite{Szeliski06}
  solves the same LP relaxation as max-sum diffusion. The same holds
  for zero-temperature versions of other recently proposed convergent
  algorithms to minimize convex free energies
  \cite{Johnson07,Globerson08,Weiss-UAI07,Hazan-UAI08}.
} \cite{Szeliski06}.

However, even when the LP relaxation is tight,
\refeq{marginals:diffusion} are a very poor approximation of
max-marginals. They are inexact even for acyclic hypergraphs.


\section{Bethe free energy and belief propagation}
\label{sec:BP}

Let the true entropy in~\refeq{conjugacy} be approximated by the {\em
  Bethe entropy\/}
\begin{equation}
H(\mu) \approx
-\<\log\mu,\mu\> + \sum_{v\in V} n_v \<\log\mu_v,\mu_v\> ,
\label{eq:H-Bethe}
\end{equation}
where $n_v$ is the number of hyperedges containing variable $v$. For
acyclic hypergraphs the Bethe entropy is equal to $H(\mu)$, otherwise
it can be non-concave and even negative on~$\Lambda$. Then
\refeq{conjugacy} reads
\begin{equation}
\max_{\mu\in \Lambda} \Big[ \<\theta-\log\mu,\mu\> + \sum_{v\in V} n_v \<\log\mu_v,\mu_v\> \Big] .
\label{eq:primal-Bethe}
\end{equation}
The negative objective of~\refeq{primal-Bethe} is the {\em Bethe free
  energy\/}. 

Next we formulate loopy belief propagation. Unlike in the
`traditional' formulation
\cite{Pearl88,Kschischang-IT-2001,Yedidia05,Wainwright08}, we identify
messages with reparameterizations, which agrees with \cite[eq.\
(2)]{Kolmogorov06} and \cite{Werner-UAI-2010}. Let the
marginals~\refeq{marginals} be approximated as
\begin{equation}
\log\mu_v(x_v) = \hat\theta_v(x_v) - \bigoplus_{y_v} \hat\theta_v(y_v), \quad
\log\mu_a(x_a) =  \hat \theta_a(x_a) - \bigoplus_{y_a} \hat \theta_a(y_a)
\label{eq:marginals:BP}
\end{equation}
where
\begin{equation}
\hat\theta_v=\theta_v , \quad
\hat \theta_a=\theta_a+\sum_{v\in a} \theta_v .
\label{eq:hattheta:BP}
\end{equation}
Note that $\mu_v$ and $\mu_a$ is the Gibbs distribution for the simple
graphical model with hypergraph $(\{v\},\emptyset)$ and $(a,\{a\})$,
respectively. This corresponds to decomposing $(V,E)$ into small
sub-hypergraphs. In general, $\mu$ fails to satisfy the local
marginalization conditions of~\refeq{local-polytope}.
Plugging~\refeq{marginals:BP} into these conditions yields
\begin{equation}
\bigoplus_{x_{a\setminus v}} \Big[ \theta_a(x_a) + \sum_{u\in a}\theta_u(x_u) \Big]
= \theta_v(x_v) + {\rm const}_{av} ,
\label{eq:fixed:BP:nocancel}
\end{equation}
which by cancelling $\theta_v(x_v)$ simplifies to
\begin{equation}
\bigoplus_{x_{a\setminus v}} \Big[ \theta_a(x_a) + \sum_{u\in a\setminus v}\theta_u(x_u) \Big]
= {\rm const}_{av} .
\label{eq:fixed:BP}
\end{equation}
Here, ${\rm const}_{av}$ is a constant independent on $x_v$.  We
define a {\em belief propagation fixed point\/} to be a
vector~$\theta$ satisfying~\refeq{fixed:BP} for all $v\in a\in E$ and
$x_v\in X_v$. The BP algorithm then tries to reparameterize $\theta$
to make it satisfy~\refeq{fixed:BP}.





As discovered by Yedidia et al.\ \cite{Yedidia05}, BP fixed
points~\refeq{fixed:BP} correspond to stationary points of
problem~\refeq{primal-Bethe} via the map~\refeq{marginals:BP}.  Heskes
\cite{Heskes-NIPS-2003} showed that every {\em stable\/} BP fixed
point is a local maximum (rather than minimum or saddle point)
of~\refeq{primal-Bethe}, but not necessarily {\em vice versa}.

\subsection{Zero-temperature limit: max-sum belief propagation}
\label{sec:BP:inf}

In the zero-temperature limit, `$\oplus$'
in~\refeq{marginals:BP}--\refeq{fixed:BP} is replaced with `$\max$'.
We assume in~\S\ref{sec:BP:inf} that this has been done.
Then, \refeq{fixed:BP} defines a fixed point of {\em max-sum belief
  propagation\/}\footnote{
  The traditional names `sum-product' and `max-product' are misnomers
  in our paper because we stated BP in the logsumexp-sum semiring
  $(\overline{\bb R},\oplus,+)$ rather than (as is usual) in the
  (isomorphic) sum-product semiring $(\bb R_+,+,\times)$. For zero
  temperature, we are then in the max-sum semiring $(\overline{\bb
    R},\max,+)$ rather than in the max-product semiring $(\bb
  R_+,\max,\times)$.
}.

According to~\S\ref{sec:var-inference}, numbers~\refeq{marginals:BP}
are approximations of max-marginals $\nu^\infty$ and the solution set
of the CSP defined by active approximate max-marginals is an
approximation of the set $\argmax_{x_V}\<\theta,\delta(x_V)\>$ of
ground states\footnote{
  Decoding an assignment from a fixed point of the loopy
  max-sum/max-product BP has been addressed in the BP literature (see
  e.g.\ \cite{Weiss-IT-2001,Wainwright08}) but never has been
  formulated as a CSP. But we believe this is a very natural
  formulation.
}. Since approximate max-marginals~\refeq{marginals:BP} are, up to
normalization, equal to numbers~\refeq{hattheta:BP}, this CSP is
defined by $\lceil\hat\theta\rceil$. This formulation is consistent
because (as is easy to verify) the value $\<\theta,\delta(x_V)\>$
is the same for all solutions $x_V$ of the CSP
$\lceil\hat\theta\rceil$.

It is well-known that the approximation of ground states obtained by
max-sum belief propagation is often poor (letting alone that the
algorithm may not converge). In our formalism, the value
$\<\theta,\delta(x_V)\>$ for the solutions $x_V$ of CSP
$\lceil\hat\theta\rceil$ are often far\footnote{
  Note that this never happens for max-sum diffusion, where solutions
  of the CSP $\lceil\theta\rceil$ are inevitably ground states.
} from $\Phi^\infty(\theta)$.  It may of course also happen that the
CSP $\lceil\hat\theta\rceil$ has no solution.
The situation is especially intriguing if the functions $\theta_a$ are
supermodular\footnote{
  For supermodular $\theta_a$, CSP $\lceil\hat\theta\rceil$ always has
  a solution. This is easy to prove: since function $\theta_a$ are
  supermodular, functions $\hat\theta_a$ are supermodular as well, and
  then the proof proceeds like the proof \cite{Werner-PAMI-2010} that
  max-sum diffusion exactly solves (permuted) supermodular problems.
}.  Then maximizing $\<\theta,\delta(x_V)\>$ is tractable but the
approximation obtained from max-sum BP can be inexact
\cite{Wainwright08}. 

On the other hand, if the approximation of ground states from max-sum
BP is good, then usually also the approximation~\refeq{marginals:BP}
of max-marginals is good. This is intuitively justified by the fact
that at a BP fixed point, the (max-)marginals are exact in every
subtree of the factor graph \cite{Wainwright03d,Wainwright04}.



\section{Direct minimization of the Bethe free energy}
\label{sec:double-loop}

Heskes \cite{Heskes-NIPS-2003,Heskes-JAIR-2006} proposed a class of
convergent algorithms to find a local minimum of Bethe and Kikuchi
free energies, based on the {\em minorize-maximize approach\/}
\cite{Hunter-2004,Sriperumbudur-NIPS-2009}.  We now describe a simple
representant of this class, which finds a local maximum of the
non-concave maximization problem~\refeq{primal-Bethe}.

Let $F(\mu)$ denote the objective of \refeq{primal-Bethe}. A family of
minorants of $F$ is constructed as
\begin{equation}
\tilde F(\mu,\tilde\mu) = \<\theta-\log\mu,\mu\> + \sum_{v\in V} n_v \<\log\tilde\mu_v,\mu_v\> ,
\label{eq:FF}
\end{equation}
where $\tilde\mu$ is a collections of variable distributions
$\tilde\mu_v$, non-negative and normalized. For any $\mu$ and
$\tilde\mu$ we have $\tilde F(\mu,\tilde\mu)\le F(\mu)$, with equality
if and only if $\tilde\mu_v=\mu_v$ for all $v\in V$.  This follows
from the well-known fact that any non-negative and normalized vectors
$\mu_v$ and $\tilde\mu_v$ satisfy $\<\log\tilde\mu_v,\mu_v\> \le
\<\log\mu_v,\mu_v\>$, which holds with equality only if
$\tilde\mu_v=\mu_v$.

The problem~\refeq{primal-Bethe} is now split into two nested problems
\begin{equation}
\max_{\tilde\mu} \max_{\mu\in\Lambda} \tilde F(\mu,\tilde\mu) .
\label{eq:maxmax}
\end{equation}
The inner problem is a concave maximization, which can be solved
optimally -- in fact, it has the form~\refeq{primal-convex}.  The
objective $\max_{\mu\in\Lambda} \tilde F(\mu,\tilde\mu)$ of the outer
problem is a non-concave function of $\tilde\mu$ and thus we can only
hope to find its local maximum.  The algorithm has two nested loops,
corresponding to the inner and outer problem. The outer iteration has
two steps:
\begin{enumerate}
\item Keeping $\tilde\mu$ fixed, find $\mu\in\Lambda$ that maximizes
$\tilde F(\mu,\tilde\mu)$.
\item For all $v\in V$, set $\tilde\mu_v\gets\mu_v$.
\end{enumerate}
Each of these two steps increases $\tilde F(\mu,\tilde\mu)$. For Step
1, this is true by definition. For Step 2, it follows from the
minorization property of $\tilde F$. The algorithm converges to a
state when $\mu$ is the maximum of $\tilde F(\mu,\tilde\mu)$ and
$\tilde\mu_v=\mu_v$, therefore $\mu$ is a local maximum
of~\refeq{primal-Bethe}.

In Step 1, $\tilde F(\mu,\tilde\mu)$ needs to be maximized over
$\mu\in\Lambda$. This can be cast in the form~\refeq{primal-convex}.
First we substitute $\log\tilde\mu=\tilde \theta$.
Note that after this substitution, the normalization condition
$\sum_{x_v}\tilde\mu_v(x_v)=1$ reads $\bigoplus_{x_v}\tilde
\theta_v(x_v)=0$.
Then
\begin{equation}
\tilde F(\mu,\tilde\mu)
= \<\theta-\log\mu,\mu\> + \sum_{v\in V} n_v \<\tilde \theta_v,\mu_v\>
= \<\hat \theta-\log\mu,\mu\>
\end{equation}
where, using that $\sum_v n_v \tilde\theta_v = \sum_a
\sum_{v\in a} \tilde\theta_v$, the vector $\hat \theta$ is given by\footnote{
  We could alternatively choose $\hat \theta$ as $\hat
  \theta_v=\theta_v+n_v\tilde \theta_v$, $\hat \theta_a=\theta_a$.
  But since~\refeq{hattheta} directly compares to~\refeq{hattheta:BP},
  the choice~\refeq{hattheta} more clearly shows the connection with
  BP fixed points.
}
\begin{equation}
\hat \theta_v=\theta_v , \quad
\hat \theta_a=\theta_a+\sum_{v\in a}\tilde \theta_v .
\label{eq:hattheta}
\end{equation}

The inner problem is dualized, which changes~\refeq{maxmax} to a
saddle-point problem. As described in~\S\ref{sec:diffusion}, the dual
is solved by reparameterizing $\hat \theta$ such that $\hat\theta$
satisfies~\refeq{fixed:diffusion} (which minimizes $U(\hat \theta)$)
and then computing $\mu$ from $\hat \theta$
using~\refeq{marginals:diffusion}. Since 
$\hat\theta_a^\alpha=\theta_a^\alpha+\sum_{v\in a}\tilde\theta_v$, we can
reparameterize~$\theta$ instead of~$\hat\theta$. The outer iteration
now reads as follows:
\begin{enumerate}
\item Reparameterize $\theta$ such that
\begin{equation}
\bigoplus_{x_{a\setminus v}} \Big[ \theta_a(x_a) + \sum_{u\in a} \tilde \theta_u(x_u) \Big] = \theta_v(x_v) .
\label{eq:step1-reparam}
\end{equation}
\item For all $v\in V$, set $\displaystyle \tilde \theta_v \gets \theta_v - \bigoplus_{x_v}\theta_v(x_v)$.
\end{enumerate}
The number
\[
U(\hat\theta) =
\sum_{v\in V} \bigoplus_{x_v} \theta_v(x_v) + \sum_{a\in E} \bigoplus_{x_a} \Big[ \theta_a(x_a)+\sum_{v\in a}\tilde \theta_v(x_v) \Big]
\]
is decreased by Step 1 and it is increased by Steps 1+2 combined. The
algorithm converges to a state when $\tilde \theta_v = \theta_v -
\bigoplus_{x_v}\theta_v(x_v)$. Then, $\theta$ is a BP fixed point.
This is indeed very obvious: since $\tilde \theta_v$ and $\theta_v$
are equal up to an additive constant, \refeq{step1-reparam} becomes
the same as~\refeq{fixed:BP:nocancel}, therefore~\refeq{fixed:BP}
holds.  If reparameterizations are represented by messages, we obtain
Algorithm~\ref{alg:double-loop}.

Let us remark that the normalization in Step~2 is not necessary, we
could just set $\tilde \theta_v \gets \theta_v$. This would not affect
convergence to a BP fixed point but $U(\hat\theta)$ would lose its
meaning and $\tilde\theta_v$ might grow unbounded.


\begin{algorithm}[htb]
\caption{Double-loop algorithm to find a BP fixed point.}
\label{alg:double-loop}
\begin{algorithmic} 
\State {\bf Initialization:} Choose any $\tilde \theta$ with $\bigoplus_{x_v}\tilde \theta_v(x_v)=0$. Choose any $\alpha$.
\Repeat \Comment{outer loop}
\Repeat \Comment{inner loop}
\For { $v\in a\in E$ and $x_v\in X_v$ such that $\theta_v(x_v)>-\infty$ }
  \State $\displaystyle \alpha_{av}(x_v) \gets \alpha_{av}(x_v) + {1\over2} \Big[ \theta^\alpha_v(x_v) - \bigoplus_{x_{a\setminus v}} \Big[ \theta^\alpha_a(x_a) + \sum_{u\in a} \tilde \theta_u(x_u) \Big] \Big]$
\EndFor
\Until{ convergence }
\State For all $v\in V$, set $\displaystyle\tilde \theta_v \gets \theta^\alpha_v - \bigoplus_{x_v}\theta^\alpha_v(x_v)$.
\label{line:outer}
\Until{ convergence }
\end{algorithmic}
\end{algorithm}

The outer loop is guaranteed to converge only if the inner loop
reaches full convergence. There is no theoretical guarantee ensuring
convergence with a finite number of inner iterations -- this
unpleasant feature is common to double-loop algorithms applied to
saddle-point problems.  However, this does not seem to be an issue in
practice.


\subsection{Zero-temperature limit}

Replacing $\theta$ with $\beta\theta$ in all the formulas and taking
the limit $\beta\to\infty$ again results in replacing `$\oplus$' with
`$\max$'. Then, Algorithm~\ref{alg:double-loop} 
converges to a max-sum belief propagation fixed point.

Though we never observed the algorithm fail to converge, its
convergence (with the inner loop run to full convergence) is only a
conjecture. The argument is that if it converges for any
$\beta<\infty$ then it is reasonable to assume that it will converge
also in the limit. But we suspect that finding a formal proof for
$\beta\to\infty$ may be difficult, especially when convergence of the
inner loop (max-sum diffusion) itself is a conjecture to date. Note
that, unlike for $\beta<\infty$, the proof cannot be based on the fact
that the value of $U^\infty(\hat\theta)$ monotonically decreases
because it often remains constant after the first several outer iterations.

\paragraph{Uniform initialization.}
Depending on the initial $\tilde\theta_v$, the algorithm can converge
to different fixed points (as we indeed observed).  Particularly
interesting is the case when the initial $\tilde\theta_v$ are all
uniform -- due to the normalization condition $\max_{x_v}\tilde
\theta_v(x_v)=0$, this means $\tilde\theta=0$.  Next we focus only on
this case.

\begin{figure}[btp]
\centering
\begin{tabular}{c@{}c@{}c}
\tiny random, 4 labels, grid &
\tiny attractive, 4 labels, grid &
\tiny repulsive, 4 labels, grid \\
\includegraphics[scale=.2]{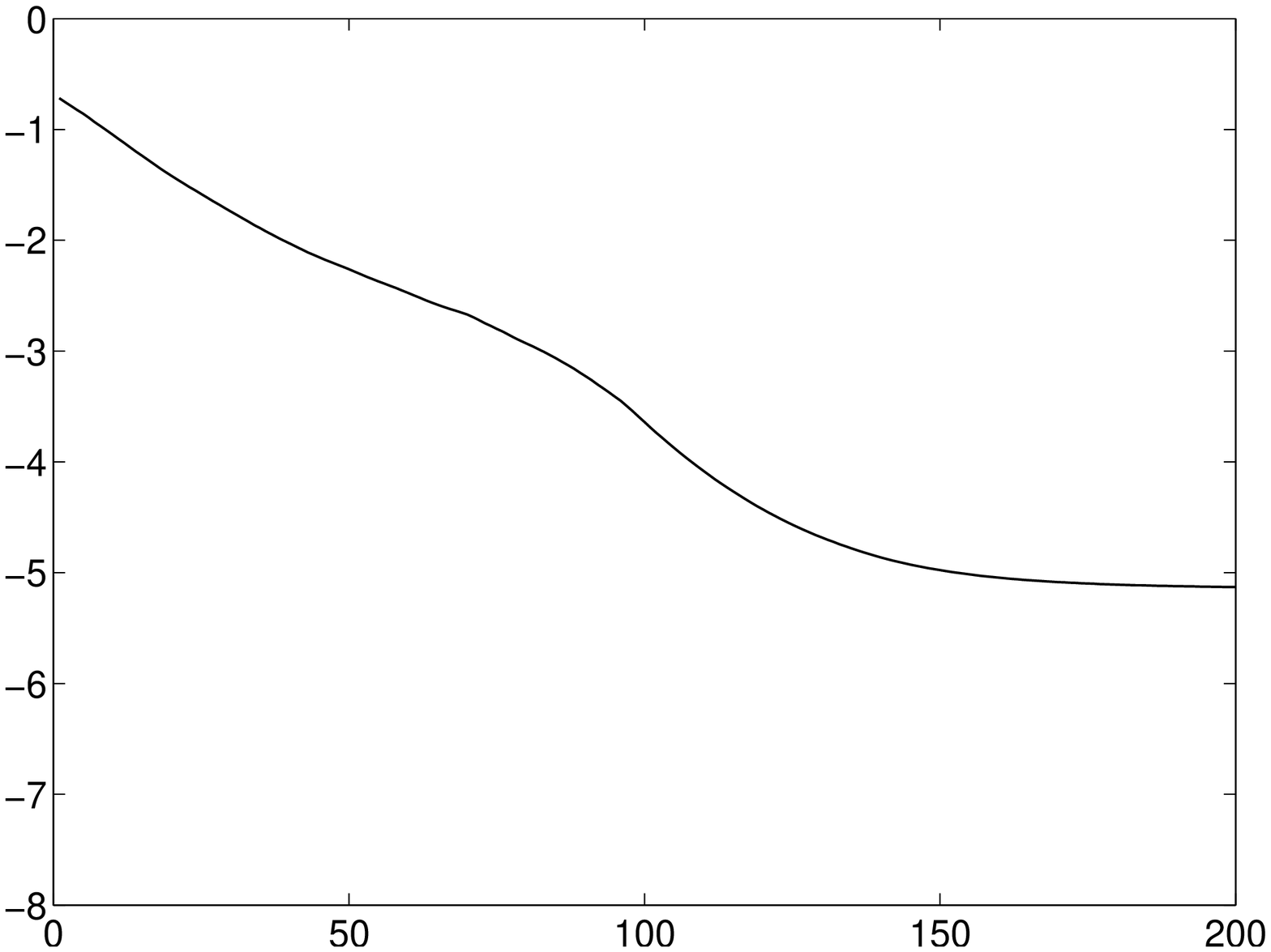} &
\includegraphics[scale=.2]{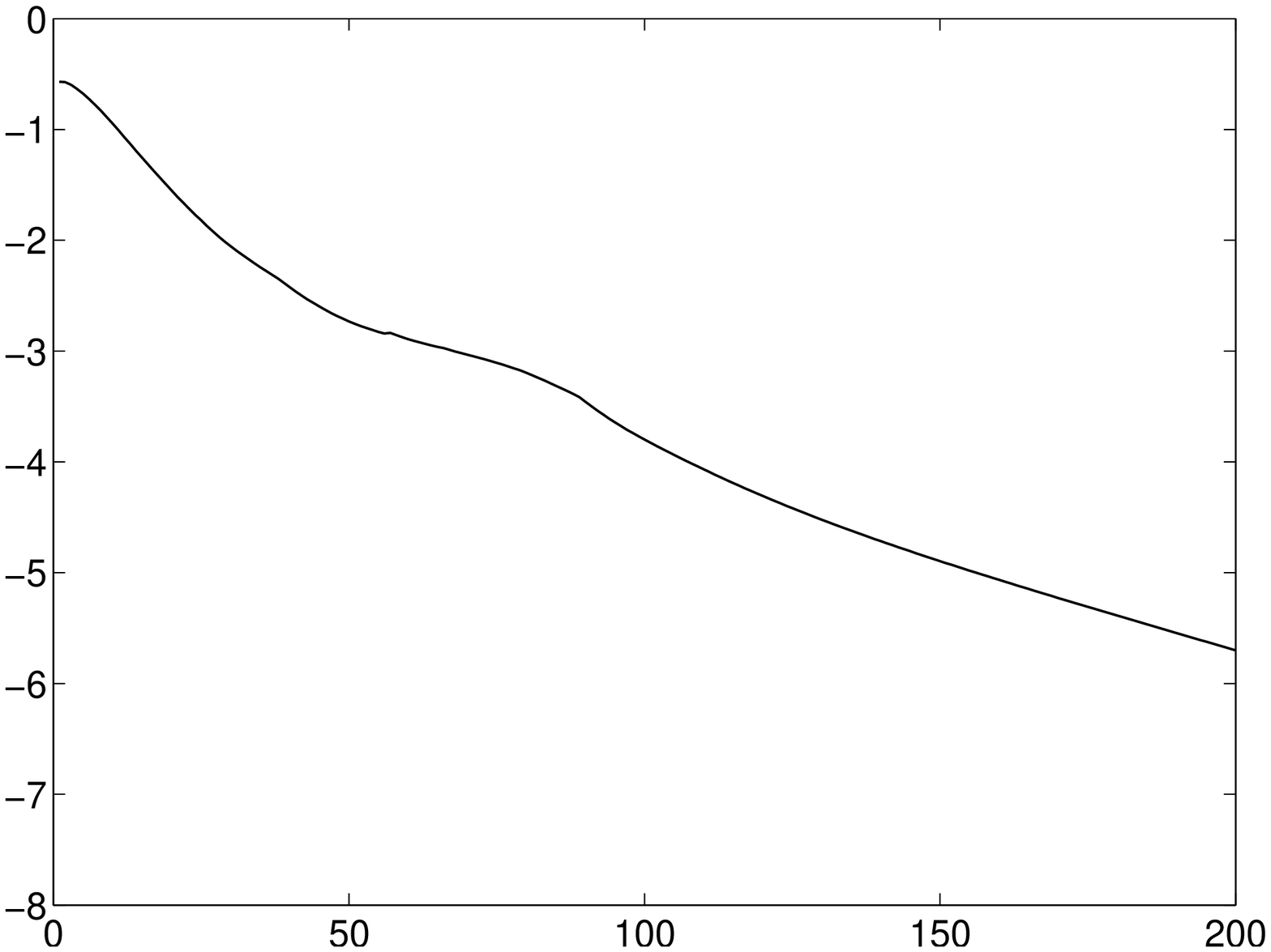} &
\includegraphics[scale=.2]{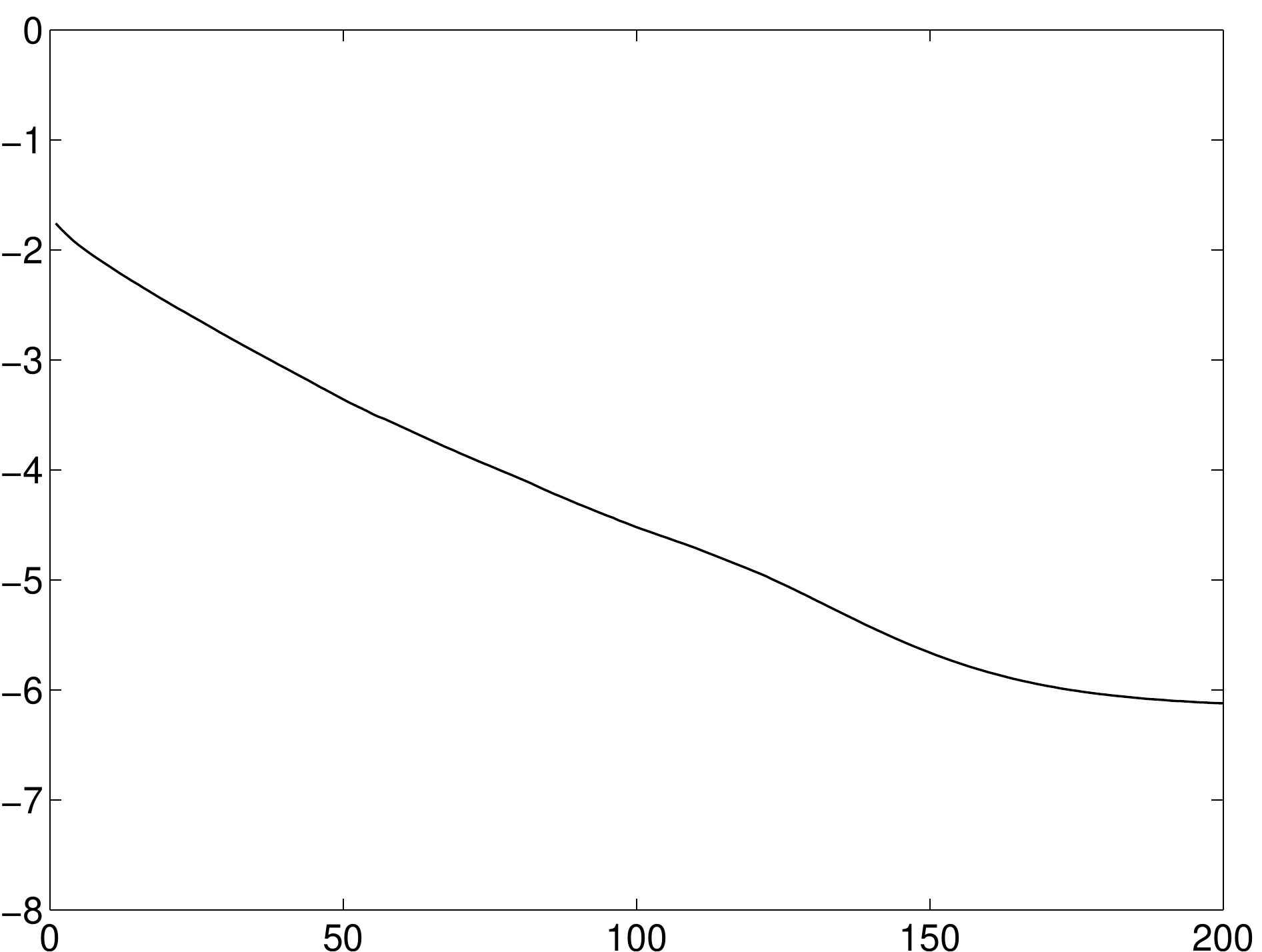} \\
\tiny repulsive, 2 labels, grid &
\tiny mixed, 4 labels, grid &
\tiny circular distance, 4 labels, grid \\
\includegraphics[scale=.2]{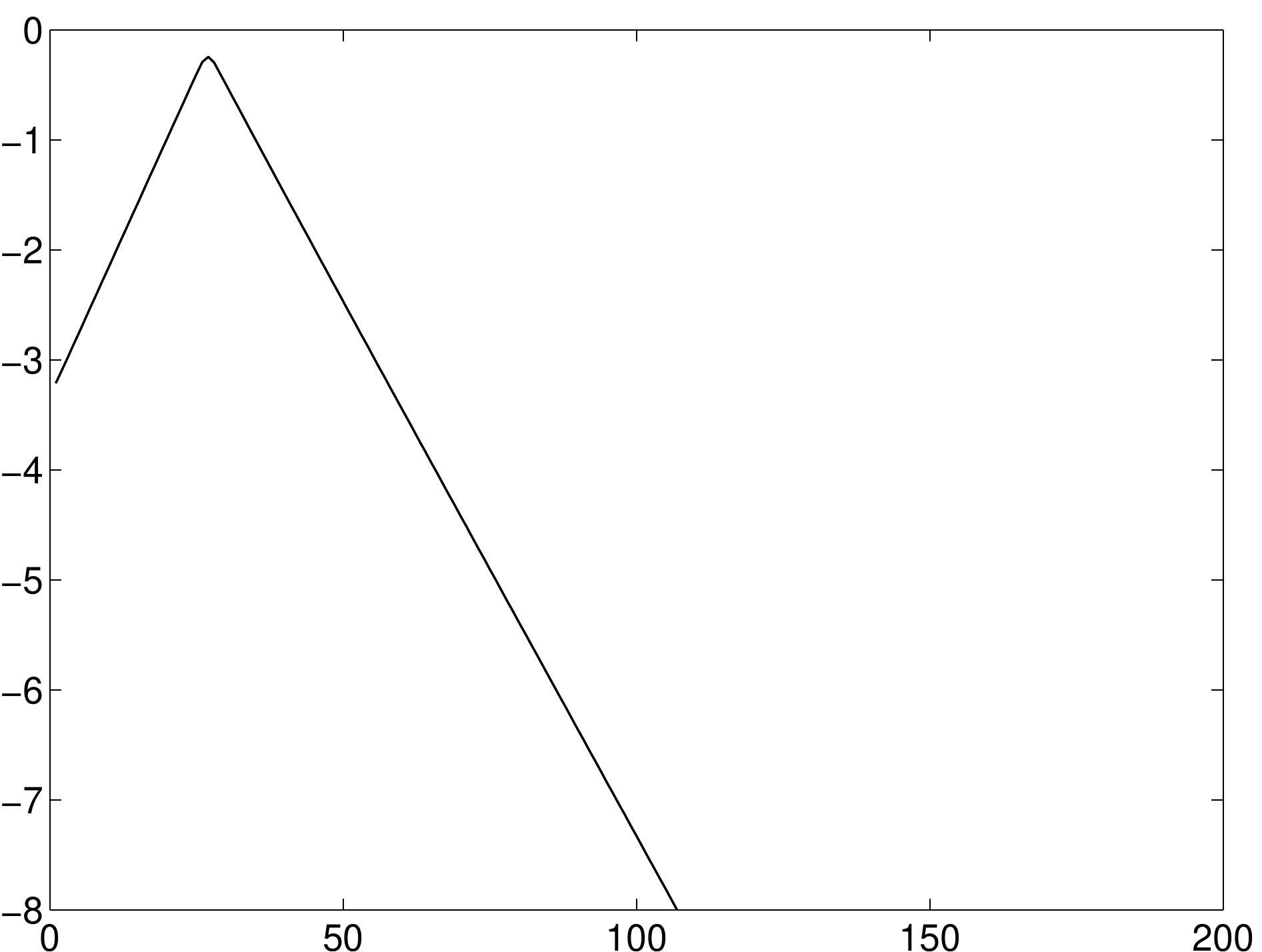} &
\includegraphics[scale=.2]{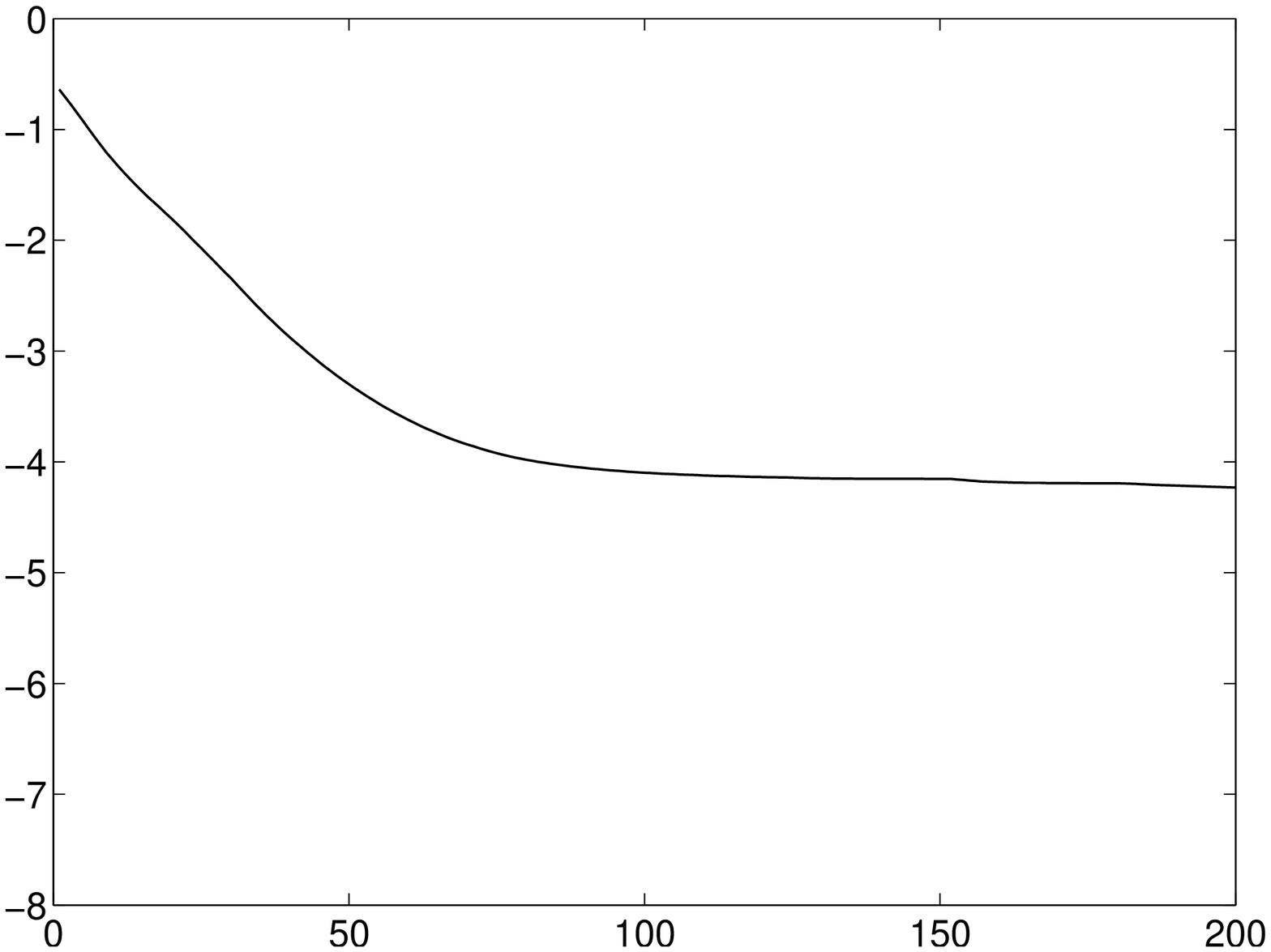} &
\includegraphics[scale=.2]{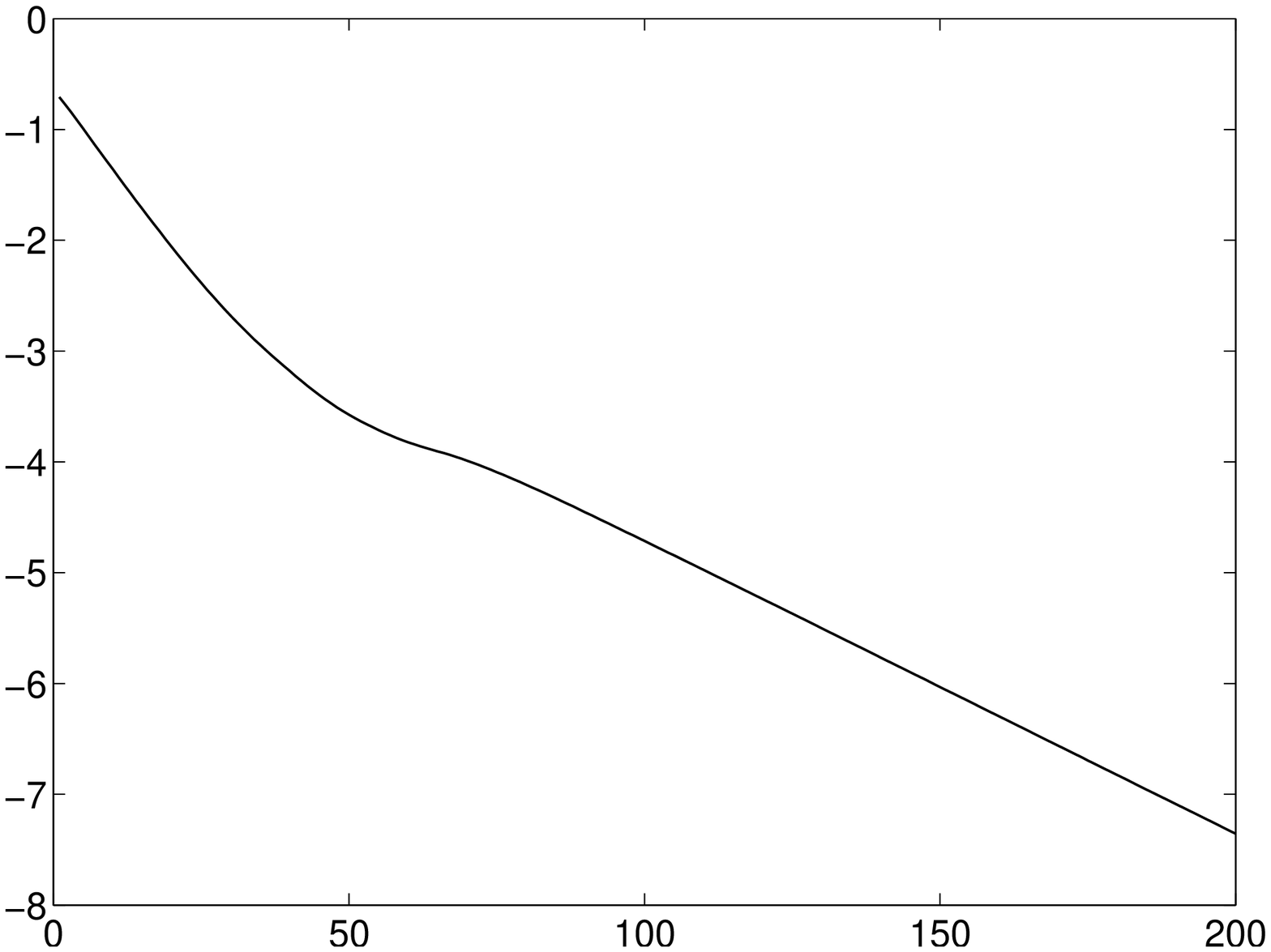} \\
\tiny random, 4 labels, complete &
\tiny attractive, 4 labels, complete &
\tiny repulsive, 4 labels, complete \\
\includegraphics[scale=.2]{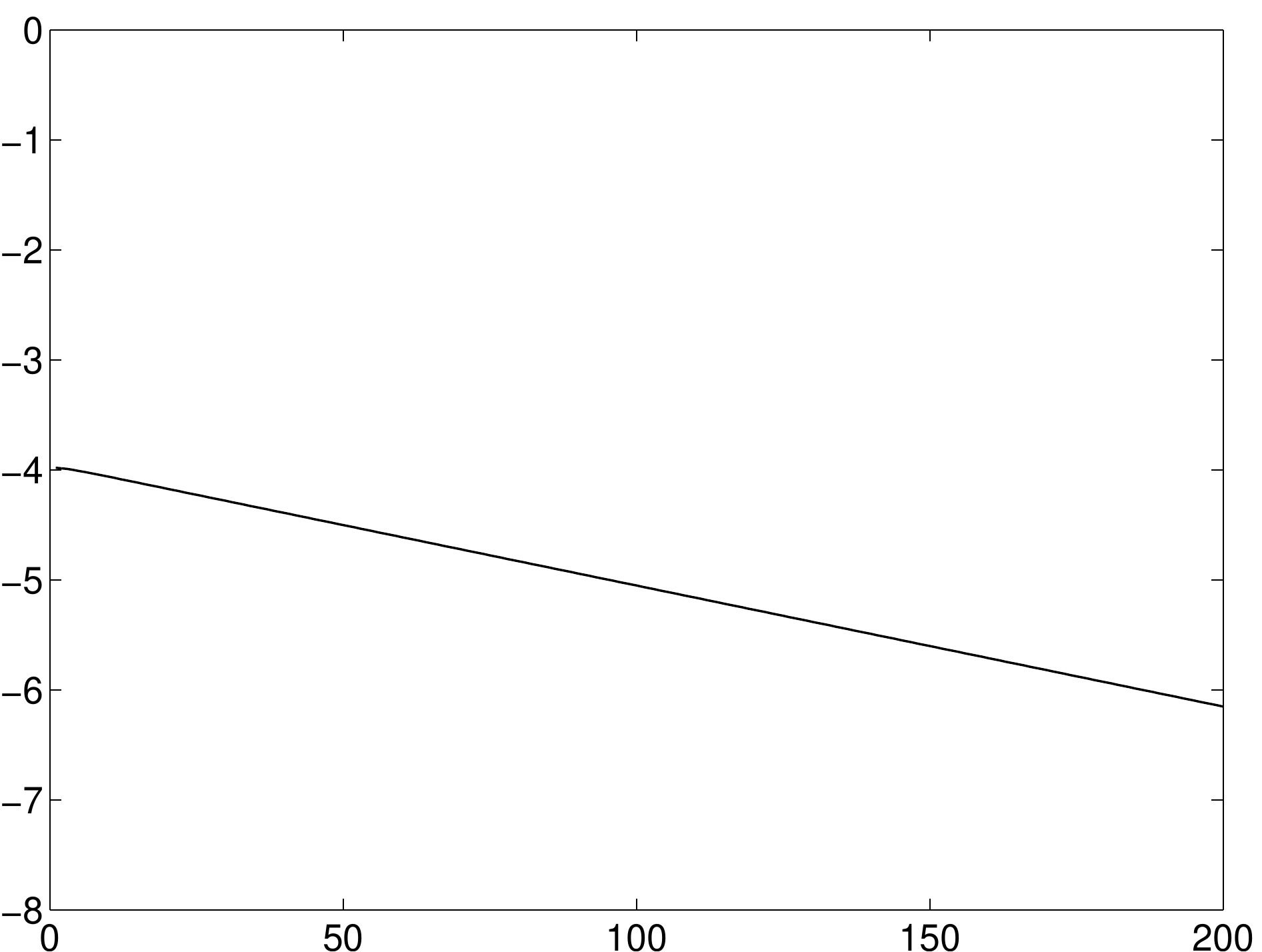} &
\includegraphics[scale=.2]{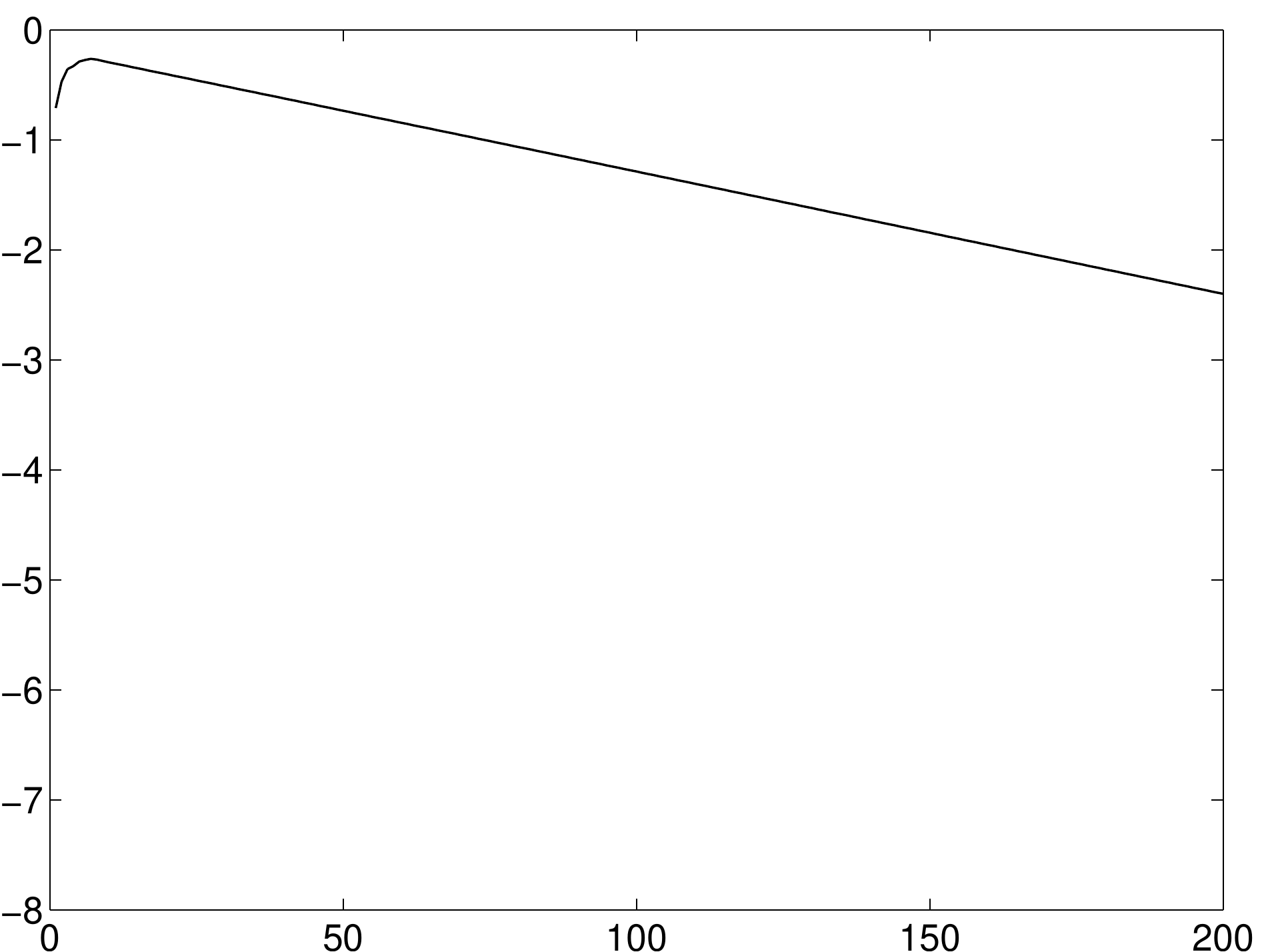} &
\includegraphics[scale=.2]{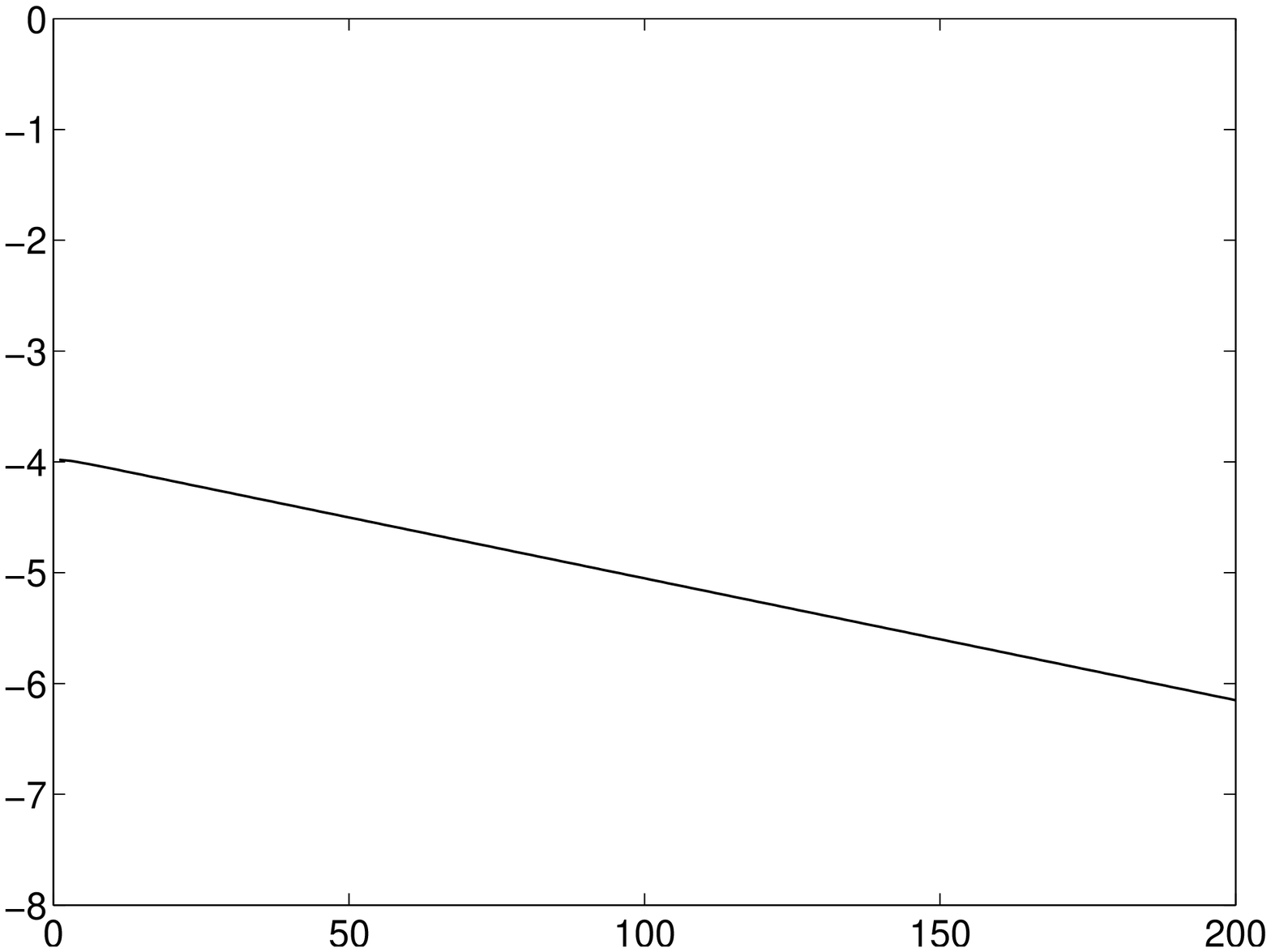}
\end{tabular}
\caption{\small Convergence of zero-temperature version of
  Algorithm~\ref{alg:double-loop} with initial $\tilde\theta_v=0$. The
  horizontal axis is the number of outer iterations, the vertical axis
  is $\log_{10}$ of average residuals to the max-sum BP fixed point
  condition~\refeq{fixed:BP:nocancel}.  The title is in the form `{\em
    type of pairwise interactions, number of labels, graph type\/}'. The
  grid graph had $20\times 20$ and the complete graph $40$
  vertices.  The unary potentials were generated randomly.}
\label{fig:resid}
\end{figure}

Figure~\ref{fig:resid} shows how the algorithm converged for different
types of pairwise interactions and different types of graph.
Occasionally (e.g., for repulsive interactions and two labels) the
residuals approached zero non-monotonically.  The inner iteration was
run to almost full convergence, however the results were not
qualitatively affected by this.

We made the following key observation:
\begin{center}
\parbox{.9\columnwidth}{ \sl If the algorithm is initialized with
  $\tilde\theta=0$ then after the first outer iteration
  $\lceil\hat\theta\rceil$ and $U^\infty(\hat\theta)$ remain
  unchanged. }
\end{center}
This observation is only empirical, currently we have neither a formal
proof nor a counterexample. It has an important consequence.  If
initially $\tilde\theta=0$, then the first outer iteration is just
Algorithm~\ref{alg:diffusion} applied to $\hat\theta=\theta$. If all
subsequent outer iterations do not change~$\lceil\hat\theta\rceil$,
then CSP~$\lceil\hat\theta\rceil$ after convergence of
Algorithm~\ref{alg:double-loop} is the same as
CSP~$\lceil\theta\rceil$ that would be obtained by running
Algorithm~\ref{alg:diffusion} on $\theta$.

Thus, the approximate ground states obtained by
Algorithm~\ref{alg:double-loop} are the same as those obtained by
Algorithm~\ref{alg:diffusion}.  However, since
Algorithm~\ref{alg:double-loop} converges to a max-sum BP fixed point,
approximate max-marginals obtained by Algorithm~\ref{alg:double-loop}
are expected to be much more accurate than those obtained by
Algorithm~\ref{alg:diffusion}.


\section{Conclusion}

We showed in~\S\ref{sec:diffusion:inf} and~\S\ref{sec:BP:inf} that the
properties of max-sum diffusion (and all MAP inference algorithms
based on LP relaxation) and max-sum belief propagation are
complementary: the former yields good approximation of ground states
but poor approximation of max-marginals, the latter {\em vice
  versa\/}. The double-loop algorithm initialized with
$\tilde\theta=0$ combines advantages of both: it yields approximate
ground states that are exact for supermodular problems and approximate
max-marginals that are exact in every sub-tree of the factor graph.

Our paper is primarily theoretical, more experiments are needed to
compare approximate max-marginals from the double loop algorithm with
ground truth.

We have not pursued another potentially interesting application of the
double-loop algorithm with non-uniform initialization
$\tilde\theta\neq0$.  It is known that max-sum BP occasionally yields
better approximate ground states than LP relaxation. This has been
observed e.g.\ for some problems on highly connected graphs
\cite{Kolmogorov06b}.  However, the max-sum BP algorithm does not
always converge, thus the convergent double loop algorithm might be
useful here.

The double-loop algorithm could be speeded up by using an inner loop
with tree updates as in e.g.\ TRW-S \cite{Kolmogorov06} rather than
edge updates as in max-sum diffusion. We believe this is possible.

\subsection*{Acknowledgments}

The work has been supported by the European Commission
project FP7-ICT-270138 and the Czech Grant Agency project P103/10/0783.
The author thanks Alexander Shekhovtsov for discussions.

\bibliographystyle{plain}
{\small \bibliography{/home/werner/publications/bib/werner}}
\end{document}